\documentclass[11pt,a4paper]{article}

\usepackage{hyperref}
\usepackage{breakurl} 
\usepackage{acl2020}

\usepackage{times}
\usepackage{appendix}
\usepackage{latexsym}
\usepackage{amsmath}
\usepackage{multirow}
\usepackage{amssymb}
\usepackage{soul}
\usepackage{bm}

\usepackage{placeins}

\usepackage{microtype}
\usepackage{xcolor}
\usepackage{graphicx}

\aclfinalcopy 

\title{GECToR -- Grammatical Error Correction: Tag, Not Rewrite}

\author{Kostiantyn Omelianchuk \qquad Vitaliy Atrasevych\thanks{~Authors  contributed equally to this work, names are given in an alphabetical order.} \quad Artem Chernodub\footnotemark[1] \qquad Oleksandr Skurzhanskyi\footnotemark[1]\\
Grammarly \\
{\tt \{firstname.lastname\}@grammarly.com} \\}

\date{}

\begin{document}
\maketitle

\begin{abstract}
{In this paper, we present a simple and efficient GEC sequence tagger using a Transformer encoder. Our system is pre-trained on synthetic data and then fine-tuned in two stages: first on errorful corpora, and second on a combination of errorful and error-free parallel corpora. We design custom token-level transformations to map input tokens to target corrections. Our best single-model/ensemble GEC tagger achieves an $F_{0.5}$ of 65.3/66.5 on CoNLL-2014 (test) and $F_{0.5}$ of 72.4/73.6 on BEA-2019 (test). Its inference speed is up to 10 times as fast as a Transformer-based seq2seq GEC system. The code and trained models are publicly available\footnote{\url{https://github.com/grammarly/gector}}.}
\end{abstract}

\section{Introduction}

Neural Machine Translation (NMT)-based approaches \cite{sennrich2016edinburgh} have become the preferred method for the task of Grammatical Error Correction (GEC)\footnote{\url{http://nlpprogress.com/english/grammatical_error_correction.html} (Accessed 1 April 2020).}. In this formulation, errorful sentences correspond to the source language, and error-free sentences correspond to the target language. Recently, Transformer-based \cite{vaswani2017attention} sequence-to-sequence (seq2seq) models have achieved state-of-the-art performance on standard GEC benchmarks \cite{bryant2019bea}. Now the focus of research has shifted more towards generating synthetic data for pretraining the Transformer-NMT-based GEC systems \cite{grundkiewicz2019neural, kiyono2019empirical}.
NMT-based GEC systems suffer from several issues which make them inconvenient for real world deployment: (i) slow inference speed, (ii) demand for large amounts of training data and  (iii) interpretability and explainability; they require additional functionality to explain corrections, e.g., grammatical error type classification \cite{bryant2017automatic}. 

In this paper, we deal with the aforementioned issues by simplifying the task from sequence generation to sequence tagging. Our GEC sequence tagging system consists of three training stages: pretraining on synthetic data, fine-tuning on an errorful parallel corpus, and finally, fine-tuning on a combination of errorful and error-free parallel corpora. 

\textbf{Related work.} LaserTagger \cite{malmi-etal-2019-encode} combines a BERT encoder with an autoregressive Transformer decoder to predict three main edit operations: keeping a token, deleting a token, and adding a phrase before a token. In contrast, in our system, the decoder is a softmax layer. PIE \cite{awasthi-etal-2019-parallel} is an iterative sequence tagging GEC system that predicts token-level edit operations. While their approach is the most similar to ours, our work differs from theirs as described in our contributions below:

\indent 1. We develop custom g-transformations: token-level edits to perform (g)rammatical error corrections. Predicting g-transformations instead of regular tokens improves the generalization of our GEC sequence tagging system. \\
\indent 2. We decompose the fine-tuning stage into two stages: fine-tuning on errorful-only sentences and further fine-tuning on a small, high-quality  dataset containing both errorful and error-free sentences. \\
\indent 3. We achieve superior performance by incorporating a pre-trained Transformer encoder in our GEC sequence tagging system. In our experiments, encoders from XLNet and RoBERTa outperform three other cutting-edge Transformer encoders (ALBERT, BERT, and GPT-2). \\

\section{Datasets}

\begin{table}
\footnotesize
\centering
\begin{tabular}{cccc}
\hline 
\textbf{Dataset}  & \textbf{\# sentences}  &\textbf{\% errorful} & \textbf{Training} \\
                 &                         &\textbf{sentences}   & \textbf{stage} \\
\hline 
PIE-synthetic & 9,000,000 & 100.0\% & I \\
Lang-8 & 947,344 & 52.5\% & II \\
NUCLE & 56,958 & 38.0\% & II \\
FCE & 34,490 & 62.4\% & II \\
W\&I+LOCNESS & 34,304 & 67.3\% & II, III  \\
\hline
\end{tabular}
\caption{\label{training-data-table} Training datasets. Training stage I is pretraining on synthetic data. Training stages II and III are for fine-tuning.}
\end{table}

Table \ref{training-data-table} describes the finer details of datasets used for different training stages.

\textbf{Synthetic data.} For pretraining stage I, we use 9M parallel sentences with synthetically generated grammatical errors \cite{awasthi-etal-2019-parallel}\footnote{\url{https://github.com/awasthiabhijeet/PIE/tree/master/errorify}}.

\textbf{Training data.} We use the following datasets for fine-tuning stages II and III: National University of Singapore Corpus of Learner English (NUCLE)\footnote{\url{https://www.comp.nus.edu.sg/~nlp/corpora.html}} \cite{dahlmeier2013building}, Lang-8 Corpus of Learner English (Lang-8)\footnote{\url{https://sites.google.com/site/naistlang8corpora}} \cite{tajiri2012tense}, FCE dataset\footnote{\url{https://ilexir.co.uk/datasets/index.html}} \cite{yannakoudakis2011new}, the publicly available part of the Cambridge Learner Corpus \cite{nicholls2003cambridge} and Write \& Improve + LOCNESS Corpus \cite{bryant2019bea}\footnote{\url{https://www.cl.cam.ac.uk/research/nl/bea2019st/data/wi+locness_v2.1.bea19.tar.gz}}.

\textbf{Evaluation data}. We report results on CoNLL-2014 test set \cite{ng2014CoNLL} evaluated by official $M^2$ scorer \cite{dahlmeier2012better}, and on BEA-2019 dev and test sets evaluated by ERRANT \cite{bryant2017automatic}. 

\section{Token-level transformations}
\label{section:transformations}

We developed custom token-level transformations $T(x_i)$ to recover the target text by applying them to the source tokens $(x_1 \ldots x_N)$. Transformations increase the coverage of grammatical error corrections for limited output vocabulary size for the most common grammatical errors, such as \textit{Spelling}, \textit{Noun Number}, \textit{Subject-Verb Agreement} and \textit{Verb Form} \cite[p. 28]{yuan2017grammatical}. 

The edit space which corresponds to our default tag vocabulary size = 5000 consists of 4971 \textit{basic transformations} (token-independent KEEP, DELETE and 1167 token-dependent APPEND, 3802 REPLACE) and 29 token-independent \textit{g-transformations}.

\textbf{Basic transformations} perform the most common token-level edit operations, such as: keep the current token unchanged (tag \textit{\$KEEP}), delete current token (tag \textit{\$DELETE}), append new token $t_{1}$ next to the current token $x_i$ (tag \textit{\$APPEND}\_$t_{1}$) or replace the current token $x_i$ with another token $t_{2}$ (tag \textit{\$REPLACE\_}$t_{2}$).

\textbf{g-transformations} perform task-specific operations such as: change the case of the current token (\textit{CASE} tags), merge the current token and the next token into a single one (\textit{MERGE} tags) and split the current token into two new tokens (\textit{SPLIT} tags). Moreover, tags from \textit{NOUN NUMBER} and \textit{VERB FORM} transformations encode grammatical properties for tokens. For instance, these transformations include conversion of singular nouns to plurals and vice versa or even change the form of regular/irregular verbs to express a different number or tense.

To obtain the transformation suffix for the \textit{VERB\_FORM} tag, we use the verb conjugation dictionary\footnote{\url{https://github.com/gutfeeling/word_forms/blob/master/word_forms/en-verbs.txt}}. For convenience, it was converted into the following format: $token_{0}\_token_{1}:tag_{0}\_tag_{1}$ (e.g., $go\_goes:VB\_VBZ$). This means that there is a transition from $word_0$ and $word_1$ to the respective tags. The transition is unidirectional, so if there exists a reverse transition, it is presented separately.

The experimental comparison of covering capabilities for our token-level transformations is in Table \ref{transforms-cover-table}. All transformation types with examples are listed in Appendix, Table \ref{transforms-examples-table}.

 \textbf{Preprocessing.} To approach the task as a sequence tagging problem we need to convert each target sentence from training/evaluation sets into a sequence of tags where each tag is mapped to a single source token. Below is a brief description of our 3-step preprocessing algorithm for color-coded sentence pair from Table  \ref{iterations-examples-table}: \\

\indent Step 1). Map each token from source sentence to subsequence of tokens from target sentence.
[\textcolor{red}{A} $\mapsto$ \textcolor{blue}{A}], [\textcolor{red}{ten} $\mapsto$ \textcolor{blue}{ten}, \textcolor{blue}{-}], [\textcolor{red}{years} $\mapsto$ \textcolor{blue}{year}, \textcolor{blue}{-}], 
[\textcolor{red}{old} $\mapsto$ \textcolor{blue}{old}], [\textcolor{red}{go} $\mapsto$ \textcolor{blue}{goes}, \textcolor{blue}{to}], [\textcolor{red}{school} $\mapsto$ \textcolor{blue}{school}, \textcolor{blue}{.}].

For this purpose, we first detect the minimal spans of tokens which define differences between source tokens $(x_1 \ldots x_N)$ and target tokens $(y_1 \ldots y_M)$. Thus, such a span is a pair of selected source tokens and corresponding target tokens. We can't use these span-based alignments, because we need to get tags on the token level. So then, for each source token $x_i$, $1 \leq i \leq N$ we search for best-fitting subsequence $\Upsilon_i = (y_{j_1} \ldots y_{j_2})$, $1 \leq j_1 \leq j_2 \leq M$ of target tokens by minimizing the modified Levenshtein distance (which takes into account that successful g-transformation is equal to zero distance).

Step 2). For each mapping in the list, find token-level transformations which convert source token to the target subsequence: [\textcolor{red}{A} $\mapsto$ \textcolor{blue}{A}]: \$KEEP, [\textcolor{red}{ten} $\mapsto$ \textcolor{blue}{ten}, \textcolor{blue}{-}]: \$KEEP, \$MERGE\_HYPHEN, [\textcolor{red}{years} $\mapsto$ \textcolor{blue}{year}, \textcolor{blue}{-}]: \$NOUN\_NUMBER\_SINGULAR, \$MERGE\_HYPHEN], 
[\textcolor{red}{old} $\mapsto$ \textcolor{blue}{old}]: \$KEEP, [\textcolor{red}{go} $\mapsto$ \textcolor{blue}{goes}, \textcolor{blue}{to}]: \$VERB\_FORM\_VB\_VBZ, \$APPEND\_\textcolor{blue}{to}, [\textcolor{red}{school} $\mapsto$ \textcolor{blue}{school}, \textcolor{blue}{.}]: \$KEEP, \$APPEND\_\{\textcolor{blue}{.}\}].

Step 3). Leave only one transformation for each source token:
\textcolor{red}{A} $\Leftrightarrow$ \$KEEP,
\textcolor{red}{ten} $\Leftrightarrow$ \$MERGE\_HYPHEN,
\textcolor{red}{years} $\Leftrightarrow$ \$NOUN\_NUMBER\_SINGULAR,
\textcolor{red}{old} $\Leftrightarrow$ \$KEEP,
\textcolor{red}{go} $\Leftrightarrow$ \$VERB\_FORM\_VB\_VBZ,
\textcolor{red}{school} $\Leftrightarrow$ \$APPEND\_\{\textcolor{blue}{.}\}. 

The iterative sequence tagging approach adds a constraint because we can use only a single tag for each token. In case of multiple transformations we take the first transformation that is not a \$KEEP tag. For more details, please, see the preprocessing script in our repository\footnote{\url{https://github.com/grammarly/gector}}.

\begin{table}
\footnotesize
\begin{tabular}{|c|c|c|}
\hline
\multirow{2}{*}{\textbf{\begin{tabular}[c]{@{}c@{}}Tag\\ vocab. size\end{tabular}}} & \multicolumn{2}{c|}{\textbf{Transformations}}                                                                                                            \\ \cline{2-3} 
                                                                                              & \textbf{\begin{tabular}[c]{@{}c@{}}Basic transf.\end{tabular}} & \textbf{\begin{tabular}[c]{@{}c@{}}All transf.\end{tabular}} \\ \hline
100                                                                                           & 60.4\%                                                                   & 79.7\%                                                                    \\ \hline
1000                                                                                          & 76.4\%                                                                   & 92.9\%                                                                    \\ \hline
5000                                                                                          & 89.5\%                                                                   & 98.1\%                                                                    \\ \hline
10000                                                                                         & 93.5\%                                                                   & 100.0\%                                                                   \\ \hline
\end{tabular}
\caption{\label{transforms-cover-table} Share of covered grammatical errors in CoNLL-2014 for basic transformations only (KEEP, DELETE, APPEND, REPLACE) and for all transformations w.r.t. tag vocabulary's size. In our work, we set the default tag vocabulary size = 5000  as a heuristical compromise between coverage and model size.}
\end{table}

\section{Tagging model architecture}
Our GEC sequence tagging model is an encoder made up of pretrained BERT-like transformer stacked with two linear layers with softmax layers on the top. We always use cased pretrained transformers in their Base configurations. Tokenization depends on the particular transformer's design: BPE \cite{sennrich2015neural} is used in RoBERTa, WordPiece \cite{schuster2012japanese} in BERT and SentencePiece \cite{kudo2018sentencepiece} in XLNet. To process the information at the token-level, we take the first subword per token from the encoder’s representation, which is then forwarded to subsequent linear layers, which are responsible for error detection and error tagging, respectively.

\section{Iterative sequence tagging approach}
To correct the text, for each input token $x_i$, $1 \leq i \leq N$ from the source sequence $(x_1 \ldots x_N)$, we predict the tag-encoded token-level transformation $T(x_i)$ described in Section \ref{section:transformations}. These predicted tag-encoded transformations are then applied to the sentence to get the modified sentence.  

Since some corrections in a sentence may depend on others, applying GEC sequence tagger only once may not be enough to fully correct the sentence. Therefore,  we use the iterative correction approach from \cite{awasthi-etal-2019-parallel}: we use the GEC sequence tagger to tag the now modified sequence, and apply the corresponding transformations on the new tags, which changes the sentence further (see an example in Table  \ref{iterations-examples-table}). Usually, the number of corrections decreases with each successive iteration, and most of the corrections are done during the first two iterations (Table \ref{iterations-stats-table}). Limiting the number of iterations speeds up the overall pipeline while trading off qualitative performance.

\begin{table}
\footnotesize
\centering
\begin{tabular}{ccc}
\hline 
\textbf{Iteration \#} & \textbf{Sentence's evolution} & \textbf{\# corr.} \\
\hline
Orig. sent & \textcolor{red}{A ten years old boy go school} & - \\
Iteration 1 & A ten\textbf {-}years old boy \textbf{goes} school & 2  \\
Iteration 2 & A ten-\textbf{year-}old boy goes \textbf{to} school & 5 \\
Iteration 3 & \textcolor{blue}{A ten-year-old boy goes to school\textbf{.}} & 6 \\
\hline
\end{tabular}
\caption{\label{iterations-examples-table} Example of iterative correction process where GEC tagging system is sequentially applied at each iteration. Cumulative number of corrections is given for each iteration. Corrections are in bold.}
\end{table}

\begin{table}
\footnotesize
\centering
\begin{tabular}{ccccc}
\hline 
\textbf{Iteration \#} & \textbf{P} & \textbf{R} & $\mathbf{F_{0.5}}$ & \textbf{\# corr.}\\ 
\hline
Iteration 1 & 72.3 & 38.6 & 61.5 & 787  \\
Iteration 2 & 73.7 & 41.1 & 63.6 & 934  \\
Iteration 3 & 74.0 & 41.5 & 64.0 & 956  \\
Iteration 4 & 73.9 & 41.5 & 64.0 & 958  \\
\hline
\end{tabular}
\caption{\label{iterations-stats-table} Cumulative number of corrections and corresponding scores on CoNLL-2014 (test) w.r.t. number of iterations for our best single model.}
\end{table}

\newpage
\section{Experiments}

\textbf{Training stages}. We have 3 training stages (details of data usage are in Table \ref{training-data-table}):
\renewcommand{\labelenumi}{\Roman{enumi}}
\begin{enumerate}
    \item Pre-training on synthetic errorful sentences as in \cite{awasthi-etal-2019-parallel}.
    \item Fine-tuning on errorful-only sentences.
    \item Fine-tuning on subset of errorful and error-free sentences as in \cite{kiyono2019empirical}.
\end{enumerate}
We found that having two fine-tuning stages with and without error-free sentences is crucial for performance (Table \ref{train-stages-xlnet-table}). 

\begin{table}
\footnotesize
\centering
\begin{tabular}{|l|c|c|c|c|c|c|}
\hline
\multicolumn{1}{|c|}{\multirow{2}{*}{\begin{tabular}[c]{@{}c@{}}\textbf{Training}\\ \textbf{stage} \#\end{tabular}}} & \multicolumn{3}{c|}{\textbf{CoNLL-2014 (test)}} & \multicolumn{3}{c|}{\textbf{BEA-2019 (dev)}} \\ \cline{2-7} 
\multicolumn{1}{|c|}{}                                                                            & \textbf{P}           & \textbf{R}           & $\mathbf{F_{0.5}}$           & \textbf{P}          & \textbf{R}          & $\mathbf{F_{0.5}}$          
\\ \hline
Stage I.  &   55.4   &  35.9   &   49.9   &   37.0   &  23.6   &   33.2   \\ 
Stage II.   &   64.4   &  46.3    &   59.7   &   46.4   &  37.9   &   44.4   \\ 
Stage III. & 66.7&  \textbf{49.9} & 62.5 & 52.6 &  \textbf{43.0}   & 50.3   \\
Inf. tweaks &\textbf{77.5}&  40.2   &\textbf{65.3}&\textbf{66.0}&  33.8   &\textbf{55.5}   \\

\hline
\end{tabular}
\caption{\label{train-stages-xlnet-table} Performance of GECToR (XLNet) after each training stage and inference tweaks.}
\end{table}

All our models were trained by Adam optimizer \cite{kingma1412adam} with default hyperparameters. Early stopping was used; stopping criteria was 3 epochs of 10K updates each without improvement. We set batch size=256 for pre-training stage I (20 epochs) and batch size=128 for fine-tuning stages II and III (2-3 epochs each). We also observed that freezing the encoder's weights for the first 2 epochs on training stages I-II and using a batch size greater than 64 improves the convergence and leads to better GEC performance.

\textbf{Encoders from pretrained transformers}. We fine-tuned BERT \cite{devlin2019bert}, RoBERTa \cite{liu2019roberta}, GPT-2 \cite{radford2019language}, XLNet \cite{yang2019xlnet}, and ALBERT \cite{lan2019albert} with the same hyperparameters setup. We also added LSTM with randomly initialized embeddings ($dim = 300$) as a baseline. As follows from Table \ref{encoder-table}, encoders from fine-tuned Transformers significantly outperform LSTMs. BERT, RoBERTa and XLNet encoders perform better than GPT-2 and ALBERT, so we used them only in our next experiments. All models were trained out-of-the-box\footnote{\url{https://huggingface.co/transformers/}} which seems to not work well for GPT-2. We hypothesize that encoders from Transformers which were pretrained as a part of the entire encoder-decoder pipeline are less useful for GECToR.

\begin{table}[ht]
\footnotesize
\centering
\begin{tabular}{c|ccc|ccc}
\hline
\multirow{2}{*}{\textbf{Encoder}} & \multicolumn{3}{c|}{\textbf{CoNLL-2014 (test)}} & \multicolumn{3}{c}{\textbf{BEA-2019 (dev)}}       \\ \cline{2-7} 
                         & \textbf{P} &  \textbf{R}  & $\mathbf{F_{0.5}}$ & \textbf{P} &  \textbf{R}  & $\mathbf{F_{0.5}}$ \\ \hline
LSTM  & 51.6 & 15.3 & 35.0 & - &- & -\\
                         \hline
ALBERT                   &   59.5   &  31.0   &   50.3   &   43.8   &  22.3   &   36.7   \\ 
BERT                     &   65.6   &  36.9   &   56.8   &   48.3   &  29.0   &   42.6   \\ 
GPT-2                    &   61.0   &  6.3    &   22.2   &   44.5   &  5.0    &   17.2   \\ 
RoBERTa                  &\textbf{67.5}&  38.3   &\textbf{58.6}&\textbf{50.3}&  30.5   &\textbf{44.5}   \\ 
XLNet                    &   64.6   &\textbf{42.6}& 58.5 &   47.1  & \textbf{34.2} &  43.8   \\ \hline
\end{tabular}
\caption{\label{encoder-table} Varying encoders from pretrained Transformers in our sequence labeling system. Training was done on data from training stage II only.}
\end{table}

\begin{table*}[h!]
\footnotesize
\centering
\begin{tabular}{lc|ccc|ccc}
\hline
\multirow{2}{*}{\textbf{GEC system}} & \multirow{2}{*}{\textbf{Ens.}}& \multicolumn{3}{c|}{\textbf{CoNLL-2014 (test)}} & \multicolumn{3}{c}{\textbf{BEA-2019 (test)}} \\ \cline{3-8} 
                        & & \textbf{P} &  \textbf{R}  & $\mathbf{F_{0.5}}$ & \textbf{P} &  \textbf{R}  & $\mathbf{F_{0.5}}$ \\ \hline
\citet{zhao2019improving}&&   67.7   &  40.6   &   59.8   &   -   &  -   &   -  \\ 
\citet{awasthi-etal-2019-parallel}&&   66.1   &  43.0    &   59.7   &   -   &  -    &   -  \\ 
\citet{kiyono2019empirical}&&   67.9  &  \textbf{44.1}   &  61.3  & 65.5 &  \textbf{59.4}   &   64.2   \\ \hline
\citet{zhao2019improving}&\checkmark&   74.1   &  36.3   &  61.3   &   -   &  -   &   -   \\ 
\citet{awasthi-etal-2019-parallel}&\checkmark&   68.3   &  43.2    &   61.2   &   -   &  -    &   -  \\ 
\citet{kiyono2019empirical}&\checkmark&   72.4  &  \textbf{46.1}   &  65.0  & 74.7 &  56.7   &   70.2    \\ 
\citet{kantor2019learning}&\checkmark&   - &  -   &  -  & 78.3 &  58.0   &   73.2   \\ \hline
GECToR (BERT) && 72.1 & 42.0     & 63.0 & 71.5  & 55.7 &  67.6  \\ 
GECToR (RoBERTa) && 73.9&  41.5   & 64.0& 77.2   & 55.1&   71.5   \\ 
GECToR (XLNet) &&\textbf{77.5}&  40.1   &\textbf{65.3}&\textbf{79.2}   & 53.9&   \textbf{72.4} \\ 
\hline
GECToR (RoBERTa + XLNet) &\checkmark& 76.6 &  42.3  & 66.0& \textbf{79.4}  & 57.2 & \textbf{73.7}   \\
GECToR (BERT + RoBERTa + XLNet) &\checkmark&\textbf{78.2}&  41.5   &\textbf{66.5}&78.9   & \textbf{58.2} &  73.6  \\ \hline
\end{tabular}
\caption{\label{results-table} Comparison of single models and ensembles. The $M^2$ score for CoNLL-2014 (test) and ERRANT for the BEA-2019 (test) are reported. In ensembles we simply average output probabilities from single models.}
\end{table*}

\textbf{Tweaking the inference}. We forced the model to perform more precise corrections by introducing two inference hyperparameters (see Appendix, Table \ref{hyperparameters-table}), hyperparameter values were found by random search on BEA-dev. 

First, we added a permanent positive \textit{confidence bias} to the probability of \$KEEP tag which is responsible for not changing the source token. 
Second, we added a sentence-level \textit{minimum error probability} threshold for the output of the error detection layer. This increased precision by trading off recall and achieved better $F_{0.5}$ scores (Table \ref{train-stages-xlnet-table}). 

Finally, our best single-model, GECToR (XLNet) achieves $F_{0.5}$ = 65.3 on CoNLL-2014 (test) and $F_{0.5}$ = 72.4 on BEA-2019 (test). Best ensemble model, GECToR (BERT + RoBERTa + XLNet) where we simply average output probabilities from 3 single models achieves $F_{0.5}$ = 66.5 on CoNLL-2014 (test) and $F_{0.5}$ = 73.6 on BEA-2019 (test), correspondingly (Table \ref{results-table}).

\textbf{Speed comparison}. We measured the model’s average inference time on NVIDIA Tesla V100 on batch size 128. For sequence tagging we don't need to predict corrections one-by-one as in autoregressive transformer decoders, so inference is naturally parallelizable and therefore runs many times faster. Our sequence tagger's inference speed is up to 10 times as fast as the state-of-the-art Transformer from \citet{zhao2019improving}, beam size=12 (Table \ref{speed-table}).

\begin{table}[ht]
\footnotesize
\centering
\begin{tabular}{l|c}
\hline
\multicolumn{1}{c|}{\textbf{GEC system}}     &  \textbf{Time (sec)} \\ \hline
Transformer-NMT, beam size = 12 &  4.35   \\
Transformer-NMT, beam size = 4   &  1.25   \\
Transformer-NMT, beam size = 1   &   0.71  \\ \hline
GECToR (XLNet), 5 iterations &  0.40           \\
GECToR (XLNet), 1 iteration &  0.20           \\ \hline
\end{tabular}
\caption{\label{speed-table} Inference time for NVIDIA Tesla V100 on CoNLL-2014 (test), single model, batch size=128.}
\end{table}

\section{Conclusions}
We show that a faster, simpler, and more efficient GEC system can be developed using a sequence tagging approach, an encoder from a pretrained Transformer, custom transformations and 3-stage training. 

Our best single-model/ensemble GEC tagger achieves an $F_{0.5}$ of 65.3/66.5 on CoNLL-2014 (test) and $F_{0.5}$ of 72.4/73.6 on BEA-2019 (test). We achieve state-of-the-art results for the GEC task with an inference speed up to 10 times as fast as Transformer-based seq2seq systems.

\section{Acknowledgements}
This research was supported by Grammarly. We thank our colleagues Vipul Raheja, Oleksiy Syvokon, Andrey Gryshchuk and our ex-colleague Maria Nadejde who provided insight and expertise that greatly helped to make this paper better. We would also like to show our gratitude to Abhijeet Awasthi and Roman Grundkiewicz for their support in providing data and answering related questions. We also thank 3 anonymous reviewers for their contribution.

\bibliography{gector}

\begin{thebibliography}{26}
\expandafter\ifx\csname natexlab\endcsname\relax\def\natexlab#1{#1}\fi

\bibitem[{Awasthi et~al.(2019)Awasthi, Sarawagi, Goyal, Ghosh, and
  Piratla}]{awasthi-etal-2019-parallel}
Abhijeet Awasthi, Sunita Sarawagi, Rasna Goyal, Sabyasachi Ghosh, and Vihari
  Piratla. 2019.
\newblock Parallel iterative edit models for local sequence transduction.
\newblock In \emph{Proceedings of the 2019 Conference on Empirical Methods in
  Natural Language Processing and the 9th International Joint Conference on
  Natural Language Processing (EMNLP-IJCNLP)}, pages 4260--4270, Hong Kong,
  China. Association for Computational Linguistics.

\bibitem[{Bryant et~al.(2019)Bryant, Felice, Andersen, and
  Briscoe}]{bryant2019bea}
Christopher Bryant, Mariano Felice, {\O}istein~E. Andersen, and Ted Briscoe.
  2019.
\newblock The {BEA}-2019 shared task on grammatical error correction.
\newblock In \emph{Proceedings of the Fourteenth Workshop on Innovative Use of
  NLP for Building Educational Applications}, pages 52--75, Florence, Italy.
  Association for Computational Linguistics.

\bibitem[{Bryant et~al.(2017)Bryant, Felice, and Briscoe}]{bryant2017automatic}
Christopher Bryant, Mariano Felice, and Ted Briscoe. 2017.
\newblock Automatic annotation and evaluation of error types for grammatical
  error correction.
\newblock In \emph{Proceedings of the 55th Annual Meeting of the Association
  for Computational Linguistics (Volume 1: Long Papers)}, pages 793--805,
  Vancouver, Canada. Association for Computational Linguistics.

\bibitem[{Dahlmeier and Ng(2012)}]{dahlmeier2012better}
Daniel Dahlmeier and Hwee~Tou Ng. 2012.
\newblock Better evaluation for grammatical error correction.
\newblock In \emph{Proceedings of the 2012 Conference of the North American
  Chapter of the Association for Computational Linguistics: Human Language
  Technologies}, pages 568--572. Association for Computational Linguistics.

\bibitem[{Dahlmeier et~al.(2013)Dahlmeier, Ng, and Wu}]{dahlmeier2013building}
Daniel Dahlmeier, Hwee~Tou Ng, and Siew~Mei Wu. 2013.
\newblock Building a large annotated corpus of learner english: The nus corpus
  of learner english.
\newblock In \emph{Proceedings of the eighth workshop on innovative use of NLP
  for building educational applications}, pages 22--31.

\bibitem[{Devlin et~al.(2019)Devlin, Chang, Lee, and
  Toutanova}]{devlin2019bert}
Jacob Devlin, Ming-Wei Chang, Kenton Lee, and Kristina Toutanova. 2019.
\newblock {BERT}: Pre-training of deep bidirectional transformers for language
  understanding.
\newblock In \emph{Proceedings of the 2019 Conference of the North {A}merican
  Chapter of the Association for Computational Linguistics: Human Language
  Technologies, Volume 1 (Long and Short Papers)}, pages 4171--4186,
  Minneapolis, Minnesota. Association for Computational Linguistics.

\bibitem[{Grundkiewicz et~al.(2019)Grundkiewicz, Junczys-Dowmunt, and
  Heafield}]{grundkiewicz2019neural}
Roman Grundkiewicz, Marcin Junczys-Dowmunt, and Kenneth Heafield. 2019.
\newblock Neural grammatical error correction systems with unsupervised
  pre-training on synthetic data.
\newblock In \emph{Proceedings of the Fourteenth Workshop on Innovative Use of
  NLP for Building Educational Applications}, pages 252--263.

\bibitem[{Kantor et~al.(2019)Kantor, Katz, Choshen, Cohen-Karlik, Liberman,
  Toledo, Menczel, and Slonim}]{kantor2019learning}
Yoav Kantor, Yoav Katz, Leshem Choshen, Edo Cohen-Karlik, Naftali Liberman,
  Assaf Toledo, Amir Menczel, and Noam Slonim. 2019.
\newblock Learning to combine grammatical error corrections.
\newblock In \emph{Proceedings of the Fourteenth Workshop on Innovative Use of
  NLP for Building Educational Applications}, pages 139--148, Florence, Italy.
  Association for Computational Linguistics.

\bibitem[{Kingma and Ba(2015)}]{kingma1412adam}
Diederik~P Kingma and Jimmy Ba. 2015.
\newblock Adam (2014), a method for stochastic optimization.
\newblock In \emph{Proceedings of the 3rd International Conference on Learning
  Representations (ICLR), arXiv preprint arXiv}, volume 1412.

\bibitem[{Kiyono et~al.(2019)Kiyono, Suzuki, Mita, Mizumoto, and
  Inui}]{kiyono2019empirical}
Shun Kiyono, Jun Suzuki, Masato Mita, Tomoya Mizumoto, and Kentaro Inui. 2019.
\newblock An empirical study of incorporating pseudo data into grammatical
  error correction.
\newblock In \emph{Proceedings of the 2019 Conference on Empirical Methods in
  Natural Language Processing and the 9th International Joint Conference on
  Natural Language Processing (EMNLP-IJCNLP)}, pages 1236--1242, Hong Kong,
  China. Association for Computational Linguistics.

\bibitem[{Kudo and Richardson(2018)}]{kudo2018sentencepiece}
Taku Kudo and John Richardson. 2018.
\newblock {S}entence{P}iece: A simple and language independent subword
  tokenizer and detokenizer for neural text processing.
\newblock In \emph{Proceedings of the 2018 Conference on Empirical Methods in
  Natural Language Processing: System Demonstrations}, pages 66--71, Brussels,
  Belgium. Association for Computational Linguistics.

\bibitem[{Lan et~al.(2019)Lan, Chen, Goodman, Gimpel, Sharma, and
  Soricut}]{lan2019albert}
Zhenzhong Lan, Mingda Chen, Sebastian Goodman, Kevin Gimpel, Piyush Sharma, and
  Radu Soricut. 2019.
\newblock Albert: A lite bert for self-supervised learning of language
  representations.
\newblock \emph{arXiv preprint arXiv:1909.11942}.

\bibitem[{Liu et~al.(2019)Liu, Ott, Goyal, Du, Joshi, Chen, Levy, Lewis,
  Zettlemoyer, and Stoyanov}]{liu2019roberta}
Yinhan Liu, Myle Ott, Naman Goyal, Jingfei Du, Mandar Joshi, Danqi Chen, Omer
  Levy, Mike Lewis, Luke Zettlemoyer, and Veselin Stoyanov. 2019.
\newblock Roberta: A robustly optimized bert pretraining approach.
\newblock \emph{arXiv preprint arXiv:1907.11692}.

\bibitem[{Malmi et~al.(2019)Malmi, Krause, Rothe, Mirylenka, and
  Severyn}]{malmi-etal-2019-encode}
Eric Malmi, Sebastian Krause, Sascha Rothe, Daniil Mirylenka, and Aliaksei
  Severyn. 2019.
\newblock Encode, tag, realize: High-precision text editing.
\newblock In \emph{Proceedings of the 2019 Conference on Empirical Methods in
  Natural Language Processing and the 9th International Joint Conference on
  Natural Language Processing (EMNLP-IJCNLP)}, pages 5054--5065, Hong Kong,
  China. Association for Computational Linguistics.

\bibitem[{Ng et~al.(2014)Ng, Wu, Briscoe, Hadiwinoto, Susanto, and
  Bryant}]{ng2014CoNLL}
Hwee~Tou Ng, Siew~Mei Wu, Ted Briscoe, Christian Hadiwinoto, Raymond~Hendy
  Susanto, and Christopher Bryant. 2014.
\newblock The conll-2014 shared task on grammatical error correction.
\newblock In \emph{Proceedings of the Eighteenth Conference on Computational
  Natural Language Learning: Shared Task}, pages 1--14.

\bibitem[{Nicholls(2003)}]{nicholls2003cambridge}
Diane Nicholls. 2003.
\newblock The cambridge learner corpus: Error coding and analysis for
  lexicography and elt.
\newblock In \emph{Proceedings of the Corpus Linguistics 2003 conference},
  volume~16, pages 572--581.

\bibitem[{Radford et~al.(2019)Radford, Wu, Child, Luan, Amodei, and
  Sutskever}]{radford2019language}
Alec Radford, Jeffrey Wu, Rewon Child, David Luan, Dario Amodei, and Ilya
  Sutskever. 2019.
\newblock Language models are unsupervised multitask learners.
\newblock \emph{OpenAI Blog}, 1(8):9.

\bibitem[{Schuster and Nakajima(2012)}]{schuster2012japanese}
Mike Schuster and Kaisuke Nakajima. 2012.
\newblock Japanese and korean voice search.
\newblock In \emph{2012 IEEE International Conference on Acoustics, Speech and
  Signal Processing (ICASSP)}, pages 5149--5152. IEEE.

\bibitem[{Sennrich et~al.(2016{\natexlab{a}})Sennrich, Haddow, and
  Birch}]{sennrich2016edinburgh}
Rico Sennrich, Barry Haddow, and Alexandra Birch. 2016{\natexlab{a}}.
\newblock {E}dinburgh neural machine translation systems for {WMT} 16.
\newblock In \emph{Proceedings of the First Conference on Machine Translation:
  Volume 2, Shared Task Papers}, pages 371--376, Berlin, Germany. Association
  for Computational Linguistics.

\bibitem[{Sennrich et~al.(2016{\natexlab{b}})Sennrich, Haddow, and
  Birch}]{sennrich2015neural}
Rico Sennrich, Barry Haddow, and Alexandra Birch. 2016{\natexlab{b}}.
\newblock Neural machine translation of rare words with subword units.
\newblock In \emph{Proceedings of the 54th Annual Meeting of the Association
  for Computational Linguistics (Volume 1: Long Papers)}, pages 1715--1725,
  Berlin, Germany. Association for Computational Linguistics.

\bibitem[{Tajiri et~al.(2012)Tajiri, Komachi, and Matsumoto}]{tajiri2012tense}
Toshikazu Tajiri, Mamoru Komachi, and Yuji Matsumoto. 2012.
\newblock Tense and aspect error correction for esl learners using global
  context.
\newblock In \emph{Proceedings of the 50th Annual Meeting of the Association
  for Computational Linguistics: Short Papers-Volume 2}, pages 198--202.
  Association for Computational Linguistics.

\bibitem[{Vaswani et~al.(2017)Vaswani, Shazeer, Parmar, Uszkoreit, Jones,
  Gomez, Kaiser, and Polosukhin}]{vaswani2017attention}
Ashish Vaswani, Noam Shazeer, Niki Parmar, Jakob Uszkoreit, Llion Jones,
  Aidan~N Gomez, {\L}ukasz Kaiser, and Illia Polosukhin. 2017.
\newblock Attention is all you need.
\newblock In \emph{Advances in neural information processing systems}, pages
  5998--6008.

\bibitem[{Yang et~al.(2019)Yang, Dai, Yang, Carbonell, Salakhutdinov, and
  Le}]{yang2019xlnet}
Zhilin Yang, Zihang Dai, Yiming Yang, Jaime Carbonell, Russ~R Salakhutdinov,
  and Quoc~V Le. 2019.
\newblock Xlnet: Generalized autoregressive pretraining for language
  understanding.
\newblock In \emph{Advances in neural information processing systems}, pages
  5754--5764.

\bibitem[{Yannakoudakis et~al.(2011)Yannakoudakis, Briscoe, and
  Medlock}]{yannakoudakis2011new}
Helen Yannakoudakis, Ted Briscoe, and Ben Medlock. 2011.
\newblock A new dataset and method for automatically grading esol texts.
\newblock In \emph{Proceedings of the 49th Annual Meeting of the Association
  for Computational Linguistics: Human Language Technologies-Volume 1}, pages
  180--189. Association for Computational Linguistics.

\bibitem[{Yuan(2017)}]{yuan2017grammatical}
Zheng Yuan. 2017.
\newblock Grammatical error correction in non-native english.
\newblock Technical report, University of Cambridge, Computer Laboratory.

\bibitem[{Zhao et~al.(2019)Zhao, Wang, Shen, Jia, and Liu}]{zhao2019improving}
Wei Zhao, Liang Wang, Kewei Shen, Ruoyu Jia, and Jingming Liu. 2019.
\newblock Improving grammatical error correction via pre-training a
  copy-augmented architecture with unlabeled data.
\newblock In \emph{Proceedings of the 2019 Conference of the North {A}merican
  Chapter of the Association for Computational Linguistics: Human Language
  Technologies, Volume 1 (Long and Short Papers)}, pages 156--165, Minneapolis,
  Minnesota. Association for Computational Linguistics.

\end{thebibliography}
\bibliographystyle{acl_natbib}

\newpage
\onecolumn
\appendix
\section{Appendix}
\newcommand{\specialcell}[2][c]{%
\begin{tabular}[#1]{@{}c@{}}#2 \end{tabular}}

\begin{table}[ht]
\scriptsize
\centering
\begin{tabular}{ccccc}
\hline 
\\
\textbf{id} & \textbf{\specialcell{Core\\ transformation}} & \textbf{\specialcell{Transformation\\suffix}}  & \textbf{Tag} & \textbf{Example} \\
\\
basic-1 & KEEP & $\mathbf{\varnothing}$ & \$KEEP &\ldots many people want to travel during the summer \ldots \\
basic-2 & DELETE & $\mathbf{\varnothing}$ & \$DELETE & \ldots not sure if you are \{\textbf{you} $\Rightarrow$ $\mathbf{\varnothing}$\} gifting \ldots \\
basic-3 & REPLACE & a & \$REPLACE\_a & \ldots the bride wears \{\textbf{the} $\Rightarrow$ \textbf{a}\} white dress \ldots \\
\ldots & \ldots & \ldots & \ldots & \ldots \\
basic-3804 & REPLACE & cause & \$REPLACE\_cause & \ldots hope it does not  \{\textbf{make} $\Rightarrow$ \textbf{cause}\} any trouble \ldots \\
basic-3805 & APPEND & for & \$APPEND\_for & \ldots he is \{\textbf{waiting} $\Rightarrow$ \textbf{waiting for}\} your reply \ldots \\
\ldots & \ldots & \ldots & \ldots & \ldots \\
basic-4971 & APPEND & know & \$APPEND\_know & \ldots I \{\textbf{don't} $\Rightarrow$ \textbf{don't know}\} which to choose\ldots
\\
\hline
g-1 & CASE & CAPITAL & \$CASE\_CAPITAL & \ldots surveillance is on the \{\textbf{internet} $\Rightarrow$ \textbf{Internet}\} \ldots \\
g-2 & CASE & CAPITAL\_1 & \$CASE\_CAPITAL\_1 & \ldots I want to buy an \{\textbf{iphone} $\Rightarrow$ \textbf{iPhone}\} \ldots \\
g-3 & CASE & LOWER & \$CASE\_LOWER & \ldots advancement in \{\textbf{Medical} $\Rightarrow$ \textbf{medical}\} technology \ldots  \\
g-4 & CASE & UPPER & \$CASE\_UPPER & \ldots the \{\textbf{it} $\Rightarrow$ \textbf{IT}\} department is concerned that\ldots  \\
g-5 & MERGE & SPACE & \$MERGE\_SPACE & \ldots insert a special kind of gene \{\textbf{in to} $\Rightarrow$ \textbf{into}\} the cell \ldots \\
g-6 & MERGE & HYPHEN & \$MERGE\_HYPHEN & \ldots and needs \{\textbf{in depth} $\Rightarrow$ \textbf{in-depth}\}  search \ldots \\
g-7 & SPLIT & HYPHEN & \$SPLIT\_HYPHEN & \ldots support us for a \{\textbf{long-run} $\Rightarrow$ \textbf{long run}\}  \ldots \\
g-8 & NOUN\_NUMBER & SINGULAR & \$NOUN\_NUMBER\_SINGULAR & \ldots a place to live for their \{\textbf{citizen} $\Rightarrow$ \textbf{citizens}\} \\
g-9 & NOUN\_NUMBER & PLURAL & \$NOUN\_NUMBER\_PLURAL & \ldots carrier of this \{\textbf{diseases} $\Rightarrow$ \textbf{disease}\} \ldots \\
g-10 & VERB FORM & VB\_VBZ & \$VERB\_FORM\_VB\_VBZ & \ldots going through this \{\textbf{make} $\Rightarrow$ \textbf{makes}\} me feel \ldots \\
g-11 & VERB FORM & VB\_VBN  &  \$VERB\_FORM\_VB\_VBN & \ldots to discuss what \{\textbf{happen} $\Rightarrow$ \textbf{happened}\} in fall \ldots \\
g-12 & VERB FORM & VB\_VBD & \$VERB\_FORM\_VB\_VBD & \ldots she sighed and \{\textbf{draw} $\Rightarrow$ \textbf{drew}\} her \ldots \\
g-13 & VERB FORM & VB\_VBG & \$VERB\_FORM\_VB\_VBG & \ldots shown success in \{\textbf{prevent} $\Rightarrow$ \textbf{preventing}\} such \ldots \\
g-14 & VERB FORM & VB\_VBZ & \$VERB\_FORM\_VB\_VBZ & \ldots a small percentage of people \{\textbf{goes} $\Rightarrow$ \textbf{go}\} by bike \ldots \\
g-15 & VERB FORM & VBZ\_VBN & \$VERB\_FORM\_VBZ\_VBN & \ldots development has \{\textbf{pushes} $\Rightarrow$ \textbf{pushed}\} countries to \ldots \\
g-16 & VERB FORM & VBZ\_VBD & \$VERB\_FORM\_VBZ\_VBD & \ldots he \{\textbf{drinks} $\Rightarrow$ \textbf{drank}\} a lot of beer last night \ldots \\
g-17 & VERB FORM & VBZ\_VBG & \$VERB\_FORM\_VBZ\_VBG & \ldots couldn't stop \{\textbf{thinks} $\Rightarrow$ \textbf{thinking}\} about it \ldots \\
g-18 & VERB FORM & VBN\_VB & \$VERB\_FORM\_VBN\_VB & \ldots going to \{\textbf{depended} $\Rightarrow$ \textbf{depend}\} on who is hiring \ldots \\
g-19 & VERB FORM & VBN\_VBZ & \$VERB\_FORM\_VBN\_VBZ & \ldots yet he goes and \{\textbf{eaten} $\Rightarrow$ \textbf{eats}\} more melons \ldots \\
g-20 & VERB FORM & VBN\_VBD & \$VERB\_FORM\_VBN\_VBD & \ldots he \{\textbf{driven} $\Rightarrow$ \textbf{drove}\} to the bus stop and \ldots \\
g-21 & VERB FORM & VBN\_VBG & \$VERB\_FORM\_VBN\_VBG & \ldots don't want you fainting and \{\textbf{broken} $\Rightarrow$ \textbf{breaking}\} \ldots \\
g-22 & VERB FORM & VBD\_VB & \$VERB\_FORM\_VBD\_VB & \ldots each of these items will
 \{\textbf{fell} $\Rightarrow$ \textbf{fall}\} in price \ldots \\
g-23 & VERB FORM & VBD\_VBZ & \$VERB\_FORM\_VBD\_VBZ & \ldots the lake \{\textbf{froze} $\Rightarrow$ \textbf{freezes}\} every year \ldots \\
g-24 & VERB FORM & VBD\_VBN & \$VERB\_FORM\_VBD\_VBN & \ldots he has been went
 \{\textbf{went} $\Rightarrow$ \textbf{gone}\} since last week \ldots \\
g-25 & VERB FORM & VBD\_VBG & \$VERB\_FORM\_VBD\_VBG & \ldots talked her into  \{\textbf{gave} $\Rightarrow$ \textbf{giving}\} me the whole day \ldots \\
g-26 & VERB FORM & VBG\_VB & \$VERB\_FORM\_VBG\_VB & \ldots free time, I just \{\textbf{enjoying} $\Rightarrow$ \textbf{enjoy}\} being outdoors \ldots \\
g-27 & VERB FORM & VBG\_VBZ & \$VERB\_FORM\_VBG\_VBZ & \ldots there still \{\textbf{existing} $\Rightarrow$ \textbf{exists}\} many inevitable factors \ldots \\
g-28 & VERB FORM & VBG\_VBN & \$VERB\_FORM\_VBG\_VBN  & \ldots people are afraid of being \{\textbf{tracking} $\Rightarrow$ \textbf{tracked}\} \ldots \\
g-29 & VERB FORM & VBG\_VBD & \$VERB\_FORM\_VBG\_VBD & \ldots there was no \{\textbf{mistook} $\Rightarrow$ \textbf{mistaking}\} his sincerity \ldots \\
\hline
\end{tabular}
\caption{\label{transforms-examples-table} List of token-level transformations (section \ref{section:transformations}). We denote a tag which defines a token-level transformation as concatenation of two parts: a \textit{core transformation} and a \textit{transformation suffix}.}
\end{table}

\begin{table}[ht]
\footnotesize
\centering
\begin{tabular}{|l|c|c|c|c|c|c|}
\hline
\multicolumn{1}{|c|}{\multirow{2}{*}{\begin{tabular}[c]{@{}c@{}}\textbf{Training}\\ \textbf{stage} \#\end{tabular}}} & \multicolumn{3}{c|}{\textbf{CoNLL-2014 (test)}} & \multicolumn{3}{c|}{\textbf{BEA-2019 (dev)}} \\ \cline{2-7} 
\multicolumn{1}{|c|}{}                                                                            & \textbf{P}           & \textbf{R}           & $\mathbf{F_{0.5}}$           & \textbf{P}          & \textbf{R}          & $\mathbf{F_{0.5}}$          
\\ \hline
Stage I.  &   57.8   &  33.0   &   50.2   &   40.8   &  22.1   &   34.9   \\ 
Stage II.   &   68.1   &  42.6    &   60.8   &   51.6   &  33.8   &   46.7   \\ 
Stage III. & 68.8&  \textbf{47.1} & 63.0 & 54.2 &  \textbf{41.0}   & 50.9   \\
Inf. tweaks &\textbf{73.9}&  41.5   &\textbf{64.0}&\textbf{62.3}&  35.1   &\textbf{54.0} \\
\hline
\end{tabular}

\caption{\label{train-stages-bert-table} Performance of GECToR (RoBERTa) after each training stage and inference tweaks. Results are given in addition to results for our best single model, GECToR (XLNet) which are given in Table \ref{train-stages-xlnet-table}.}

\end{table}

\begin{table}[!hb]
\footnotesize
\centering
\begin{tabular}{ccc}
\hline
\textbf{System name} & \textbf{Confidence bias} & \textbf{Minimum error probability} \\ 
GECToR (BERT) & 0.10 & 0.41  \\ 
GECToR (RoBERTa) & 0.20 & 0.50  \\ 
GECToR (XLNet) & 0.35 & 0.66 \\ 
GECToR (RoBERTa + XLNet) & 0.24 & 0.45  \\ 
GECToR (BERT + RoBERTa + XLNet) & 0.16 & 0.40  \\ 
\hline
\end{tabular}
\caption{\label{hyperparameters-table} Inference tweaking values which were found by random search on BEA-dev.} 
\end{table}

\end{document}